%% file: icml2021.tex
\documentclass{article}

\usepackage{microtype}
\usepackage{graphicx}
\usepackage{multirow}
\usepackage[normalem]{ulem}
\useunder{\uline}{\ul}{}
\usepackage{subfigure}
\usepackage{textcomp}
\usepackage{booktabs} 

\usepackage{array}

\makeatletter
\newcommand{\thickhline}{%
    \noalign {\ifnum 0=`}\fi \hrule height 1pt
    \futurelet \reserved@a \@xhline
}
\newcolumntype{"}{@{\hskip\tabcolsep\vrule width 1pt\hskip\tabcolsep}}
\makeatother

\usepackage{hyperref}

\usepackage[preprint]{icml2021}
\usepackage{amsfonts,amsmath,amssymb,amsthm}
\usepackage{bm}

\icmltitlerunning{Evaluating Multi-label Classifiers with Noisy Labels}
\begin{document}

\twocolumn[
%


\icmltitle{Evaluating Multi-label Classifiers with Noisy Labels}




\begin{icmlauthorlist}
\icmlauthor{Wenting Zhao}{cornell}
\icmlauthor{Carla Gomes}{cornell}
\end{icmlauthorlist}

\icmlaffiliation{cornell}{Department of Computer Science, Cornell University, Ithaca, New York, USA}

\icmlcorrespondingauthor{Wenting Zhao}{wzhao@cs.cornell.edu}

\icmlkeywords{Machine Learning, ICML}

\vskip 0.3in
]



\printAffiliationsAndNotice{}  

\nocite{langley00}

\input{abstract}
\input{introduction}
\input{noise}
\input{background}
\input{method}
\input{experiment}
\input{related}
\input{conclusion}

\clearpage
\newpage
\bibliography{references}
\bibliographystyle{icml2021}

\end{document}

%% file: abstract.tex
\begin{abstract}
    Multi-label classification (MLC) is a generalization of standard classification where multiple labels may be assigned to a given sample.
    In the real world, it is more common to deal with noisy datasets than clean datasets, given how modern datasets are labeled by a large group of annotators on crowdsourcing platforms, but little attention has been given to evaluating multi-label classifiers with noisy labels.
    Exploiting label correlations now becomes a standard component of a multi-label classifier to achieve competitive performance. 
    However, this component makes the classifier more prone to poor generalization -- it overfits labels as well as label dependencies.
    We identify three common real-world label noise scenarios and show how previous approaches perform poorly with noisy labels.
    To address this issue, we present a Context-Based Multi-Label Classifier (CbMLC) that effectively handles noisy labels when learning label dependencies, without requiring additional supervision.
    We compare CbMLC against other domain-specific state-of-the-art models on a variety of datasets, under both the clean and the noisy settings.
    We show CbMLC yields substantial improvements over the previous methods in most cases.
    
\end{abstract}

%% file: introduction.tex
\section{Introduction}

Unlike a traditional single-output learning problem where we assign one label to a given sample, multi-label classification (MLC) aims to predict a set of target objects simultaneously.
MLC has been an active research topic in domains including computer vision~\cite{wang2016cnn}, natural language processing~\cite{nam2014large}, and recommendation systems~\cite{liu2017deep}.
Notable applications are assigning tags to online images, identifying topics for web articles, and generating product recommendations.
Developing an effective MLC technique can lead to improvements in many real-world problems.

One challenge MLC faces is the introduction of complex noise in the annotation process.
A common way to collect labels is to use crowdsourcing platforms such as Amazon Mechanical Turk (MTurk).
Because a large-scale dataset could be labeled by potentially thousands of different human annotators, the labels they provide could naturally be in varying qualities~\cite{snow2008cheap,ipeirotis2010quality,raykar2010learning,yuen2011survey}; the interface a crowdsourcing platform provides can also introduce bias~\cite{li2017webvision}.
We refer readers to \citet{openimages} for an example of a data collection process for a dataset of millions of samples, where 26.6\% labels are false positive~\cite{veit2017learning}!
Another source of labels is citizen science projects.
For example, in the \emph{eBird} project~\cite{sullivan2009ebird}, amateurs observe and record birds at locations they have been to, but one may misidentify a species, and it is difficult to differentiate between one species not being a habitant of a place and the species simply not yet being observed at the place.
To develop an accurate MLC method, it is important to consider how label noise is introduced and how it may affect performance.

\begin{figure}[t]
    \centering
    \includegraphics[width=0.45\textwidth]{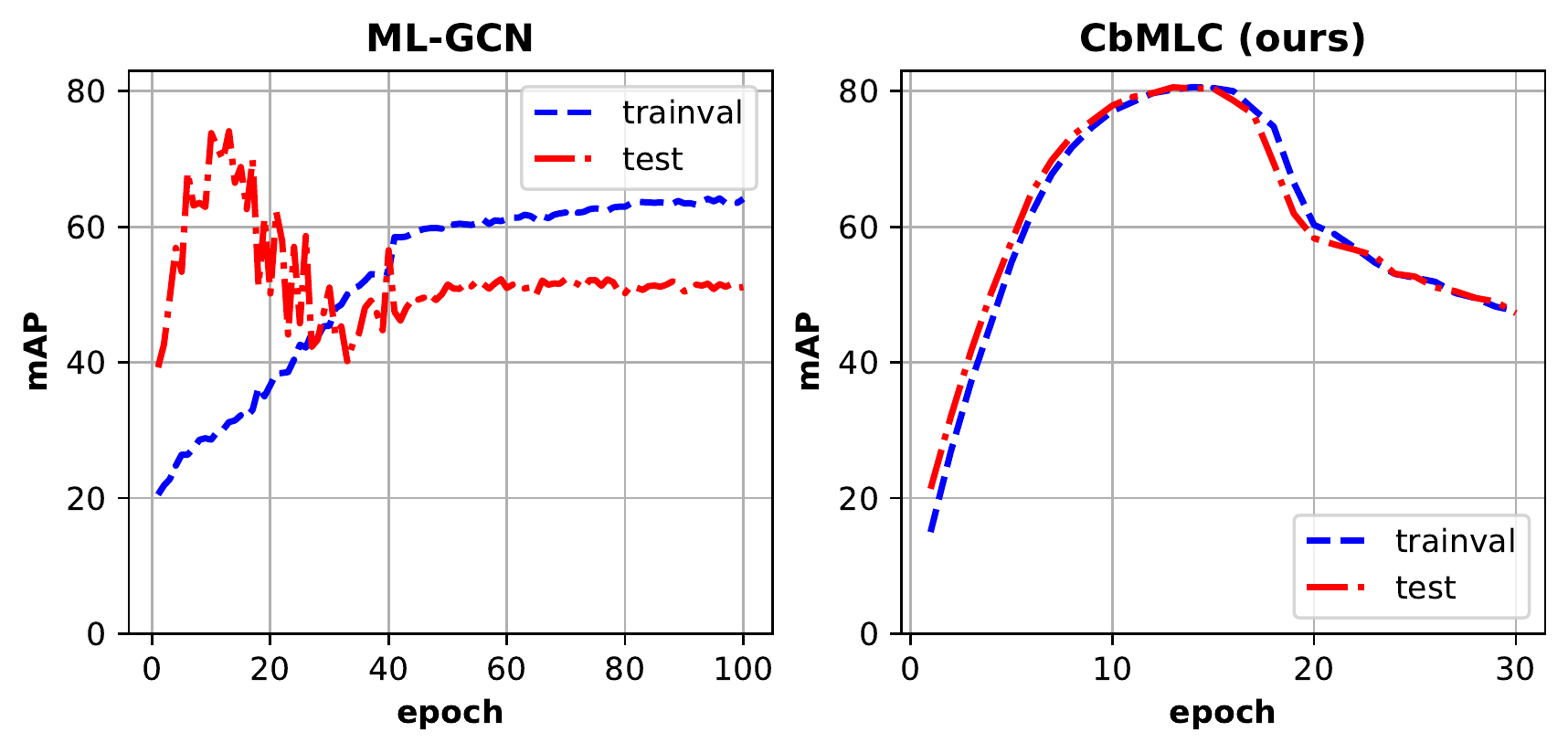}
    \caption{Mean average precision (mAP) curves on Pascal VOC produced by ML-GCN and our method, CbMLC. We inject different types of asymmetric noise on the trainval set and test with the clean test set.
    ML-GCN quickly overfits the trainval set in the first ten epochs, whereas CbMLC generalizes well on the test set as the trainval accuracy grows.}
    \label{fig:generalization}
\end{figure}

Considering label noise for the single-label setting is straightforward, as each instance only has one positive label.
That is, a mislabeled instance has exactly two classes that are flipped.
However, label noise under the multi-label setting is more complicated, as there is an unknown number of positive labels associated to an instance.
In other words, the number of mislabeled classes is arbitrary.
Despite the prevalence of label noise in MLC, little attention has been given to evaluate MLC with noisy labels.
Among the several works~\cite{li2021towards,bai2020disentangled,yao2018deep} that consider noisy labels, they only evaluate with uniform noise that is symmetric on positive and negative labels.
In the real world, positive labels in MLC are very sparse, and therefore it is unreasonable to assume symmetric noise across positive and negative labels.
We are in need of a way to model noisy labels in MLC that is closer to real-world scenarios.

Recent successes in MLC are the outcome of leveraging label correlations via graph neural networks (GNNs)~\cite{scarselli2008graph} which model how labels interact with each other~\cite{lanchantin2019neural,chen2019multi,wang2020multi}, or via label covariance matrices~\cite{chen2017deep, bai2020disentangled, tang2018multi}.
However, these methods are particularly prone to errors.
For example, ML-GCN~\cite{chen2019multi}, a state-of-the-art multi-label classification model for image recognition, pre-computes a label correlation matrix purely based on the label co-occurrences in the training set, and uses the matrix to guide information propagation among label nodes during training.
Given the over-parameterized nature of deep neural networks, it is easy to overfit label noise, leading to poor generalization~\cite{45820}.
In Figure~\ref{fig:generalization}, we show the mean averaged precision (mAP) curves for ML-GCN and our proposed method, CbMLC, on a corrupted version of \emph{Pascal VOC}, where we inject different types of asymmetric noise in the trainval set.
ML-GCN quickly overfits the noisy trainval set, and test accuracy starts dropping in the first 10 epochs.
However, using the proposed regularization technique, CbMLC substantially shrinks the generalization gap.


\textbf{Our contributions} are
(1) We identify three real-world noise scenarios and show how state-of-the-art MLC techniques perform poorly under these scenarios. We test with both symmetric and asymmetric noise types, and combined noise types with varying ratios on different instances.
(2) We propose a Context-based Multi-label Classifier (CbMLC), an end-to-end framework that leverages large pretrained word embeddings to perform context-based regularization for multi-label classification to avoid overfitting.
It employs an encoder-to-decoder network architecture with an asymmetric loss function, a domain-specific encoder to encode features into a latent vector, and a message passing decoder operating on a label graph to learn label dependencies.
This proposed framework not only benefits from employing a graph neural network to learn label correlations, but also alleviates the overfitting issue \emph{without requiring access to any clean labels}.
(3) We conduct extensive experiments on public datasets in a variety of application domains including computer vision (CV) and natural language processing (NLP). On clean datasets, our framework outperforms (or is comparable to) the state-of-the-art methods. On datasets with noise, CbMLC produces substantial improvements over previous methods in most cases.
We provide an ablation study that shows CbMLC learns better label semantics guided by  the context-based regularization.

%% file: noise.tex
\section{Modeling Real-World Noise in MLC}
\label{sec:noise}
Though there is a vast amount of research focusing on deep learning for single-label classification~\cite{song2020learning}, little attention has been drawn to the multi-label setting.
In particular, few works have examined how to model noise in a learning algorithm so that it is closer to real-world scenarios, nor is it a common practice for researchers to evaluate their MLC methods with noisy labels.

To identify possible noise types, we first discuss how noise is generated.
Due to ever-increasing dataset sizes, it becomes impossible to have human experts to annotate all available data; hence, we turn to crowd-sourcing platforms~\cite{yuen2011survey}.
A problem these platforms suffer from is that the reliability of individual workers varies, and there are individuals who are careless and give very noisy responses~\cite{snow2008cheap}.
Other automatic methods such as $k$-NN induction are used~\cite{10.1145/1646396.1646452}. 
Additionally, in citizen science projects, even people with more expertise may also make mistakes due to various reasons such as environments and perceptual errors~\cite{mcnicol1972statistical}.


We consider label noise from a statistical point of view, inspired by the noise taxonomy for the single-label setting~\citet{frenay2013classification}.
To model the label noise process, we introduce three random variables: $Y$ is the true label for one class being either 0 or 1, $\Tilde{Y}$ is the annotated label, and $E$ is a binary variable indicating whether an error occurred (i.e., $Y \neq \Tilde{Y}$).
We focus on two statistical models of label noise.
The first model is the \emph{noisy completely at random (NCAR)} model, where the occurrence of an error $E$ is independent of $Y$.
Formally, the error rate, or the probability of a true label different from the observed label is, $p_e = P(E = 1) = P(Y \neq \Tilde{Y})$.
In the case of MLC, we assume a biased coin is flipped independently for every class label of an instance to decide whether the observed label is correct or not.
Therefore, if the error rate is $p_e$ in a labeling process, an annotated instance will in expectation have $|\mathcal{Y}| \cdot p_e$ corrupted labels, where $|\mathcal{Y}|$ is the number of labels.
The second model is the \emph{noisy at random (NAR)} model, where the probability of an error $E$ depends on $Y$.
In other words, this model can represent anisotropic noise varying from classes to classes.
Formally, the error rate for a label in an instance is $p_e = P(E = 1) = P(Y=1) \cdot P(E = 1|Y=1)+P(Y=0)\cdot P(E = 1|Y=0)$.
We note these two noise models are independent of features.
While features could also impact how noise is introduced because mislabeled samples are likely similar to examples of another class~\cite{beigman2009annotator}, or have labels with low density~\cite{denoeux2000neural}, due to its complexity we leave it to future work.

Based on the two noise models, we evaluate MLC methods with the following noise types that are likely to occur in the real-world annotation processes:
\begin{enumerate}
    \item Under NCAR, we consider an example with $x\%$ labels uniformly flipped. This is by far the most widely evaluated noise type in the existing MLC literature~\cite{li2021towards,bai2020disentangled}.
    \item Under NAR, we consider an example with $x\%$ positive labels uniformly flipped. It is common to see this type of noise in citizen science projects~\cite{sullivan2009ebird}. In general, it is more often to label positive classes as negative in MLC~\cite{ben2020asymmetric}.
    \item Under NAR, we consider an example with only one positive label that is selected from all positive labels. This is a standard assumption made when collecting data from the web~\cite{li2017webvision}.
    In fact, because of ignoring the intrinsic multi-label nature of the images, the classification performance on ImageNet~\cite{russakovsky2015imagenet} produced by state-of-the-art classifiers is under-estimated~\cite{stock2018convnets}.
\end{enumerate}

We thoroughly evaluate state-of-the-art MLC methods and CbMLC with the three noise types in Section~\ref{sec:exp}.
We note that in the real world, it is rare to have uniform noise or to have only one noise type in a dataset, and thus we test the MLC algorithms with a single noise type as well as combined noise types with varying ratios.

%% file: background.tex
\section{Background}
\label{sec:background}
\paragraph{Notations.}
Let $\mathcal D$ denote the dataset $\{(\bm{x}_i,\bm{y}_i)\}_{i=1}^N$, where $\bm{x}_i \in \mathbb R^S$ is an input, and $\bm{y}_i \in \{0,1\}^L$ is an output associated with sample $i$.
Input $\bm{x}_i$ can be either an ordered or an unordered set, and output $\bm{y}_i$ has $L$ binary labels with 1 indicating the presence of a label and 0 otherwise.
We note that in the noisy learning setting, $\bm{y}_i$ could be flipped from the actual truth value.
Our goal is to make few mistakes on unseen data even when there are corrupted labels.

\paragraph{Word Embeddings.}
Word embedding is a language modeling technique where words from the vocabulary are mapped to vectors of real values.
They are often trained from very large online text corpora in an unsupervised manner.
Common word embeddings include GloVe~\cite{pennington2014glove}, GoogleNews~\cite{mikolov2013efficient}, FastText~\cite{joulin2016fasttext}.
we will use GloVe, as it features a nearest-neighbors property, where words that are semantically similar will have smaller Euclidean distances.

\paragraph{Attention-based Message Passing Neural Networks (MPNN).}
\emph{Self-attention}, or \emph{intra-attention}, is a mechanism that learns to assign different importance weights to each part of a larger piece to focus on more important portions~\cite{vaswani2017attention}.
Let $G=(V,E)$ be a directed graph where $V$ is the node set and $E$ is the edge set.
We assume $G$ to be a complete graph. 
To define self-attention on a graph in which each node attends to every other nodes and all edges carry attention weights, attention weight $a_{lj}^t$ for a node pair $(\bm{v}_l, \bm{v}_j)$ is computed as:
\begin{align}
    \label{eq:alpha}
    e^t_{lj} = a(\bm{v}^t_l,\bm{v}^t_j) = \frac{(\mathbf{W}^q\bm{v}^t_l)^{\top}
    (\mathbf{W}^u\bm{v}^t_j)}{\sqrt{\smash[b]d}} \\
    \alpha^t_{lj} = \textrm{softmax}_j(e^t_{lj}) =  \frac{\textrm{exp}(e^t_{lj})}{\sum_{k \in \mathcal{N}(l)}{\textrm{exp}(e^t_{lk})}},
\end{align}
where $a(\cdot)$ is a dot product with node-wise linear transformations $\mathbf{W}^q \in \mathbb R^{d \times d}$ on node $\bm{v}_l^t$ and $\mathbf{W}^u \in \mathbb R^{d \times d}$ on node $\bm{v}_j^t$, scaled by $\sqrt{d}$;  $e_{lj}^t$ represents the raw importance of label $j$ to label $l$ and is normalized by $\textrm{softmax}(\cdot)$ to obtain $\alpha^t_{lj}$.

We briefly introduce LaMP, which applies self-attention to learn label interactions for MLC~\cite{lanchantin2019neural}.
Each $v \in V$ associates with a label.
To predict whether a label is present, LaMP collects states from its neighbors based on the computed attention weights and makes an update on the current state of the node.
If an attention weight is large from one label to another, it suggests the one label has a strong dependency on the other label.

We describe this process formally.
In the $t$-th self-attention layer, each label is represented by a node $\bm{v}_l^t \in \mathbb R^d$.
LaMP first generates the message $\bm{m}_l^t$ of $\bm{v}_l^t$:
\begin{gather}
    \label{eq:m_attention}
    M_{\textrm{atn}}(\bm{v}^t_l,\bm{v}^t_j) = \alpha^t_{lj} \mathbf{W}^v \bm{v}^t_j,\\
    \bm{m}^t_l = \bm{v}^t_l + 
    \sum_{j\in \mathcal{N}(l)} M_{\textrm{atn}}(\bm{v}^t_l,\bm{v}^t_j), \label{eq:mt}
\end{gather}
where $\mathbf{W}^v \in \mathbb R^{d \times d}$ is a node-wise linear transformation.

After going through the $t$-th feed forward layer $U^t$, we obtain $\bm{v}_l^{t+1}$ in the $(t+1)$-th self-attention layer as:
\begin{align}
    {\bm{v}}^{t+1}_l &= \bm{m}^t_l + U^{t}(\bm{m}^{t}_l; \bm{W}).
    \label{eq:u_attention}
\end{align}
Finally, we can feed this into a multi-layer perceptron (MLP) to obtain a probability for label $\bm{v}$.
In later sections, we will refer to this module as an attention-based MPNN.

\paragraph{Asymmetric Loss Functions.}
The most widely-used loss function for MLC is binary cross entropy (BCE).
For each training sample, BCE is defined as follows:
\begin{equation}
    \label{eq:bce_loss}
    \resizebox{0.4\textwidth}{!}{$\textrm{BCE}(\bm{y},\hat{\bm{y}}) = \frac{1}{L}\sum_{l=0}^{L-1} -(y_l\log(\hat{y}_l)+(1-y_l)\log(1-\hat{y}_l))$}
\end{equation}
where $\bm{y}$ is the annotated label vector and $\hat{\bm{y}}$ is the predicted label vector.
The label class is indexed by $l$.
An issue BCE introduces is the penalty for misclassifying every label class is equal, and thus the loss reduced by easy negative labels may outweigh the penalty from the rare positive samples.
Therefore, a loss to decouple the focusing levels of the positive and negative classes would be useful.
\citeauthor{ben2020asymmetric} propose asymmetric BCE that enables this flexibility:
\begin{equation}
    \label{eq:asl}
    \resizebox{0.45\textwidth}{!}{$\textrm{ASL}(\bm{y},\hat{\bm{y}}) = \frac{1}{L}\sum_{l=0}^{L-1} -y_l(1-\hat{y}_l)^{\gamma^+}\log(\hat{y}_l) \\
    -(1-y_l)\hat{y}_l^{\gamma^-}\log(1-\hat{y}_l))$}
\end{equation}
where $\gamma^+$, $\gamma^-$ are the focusing parameters for positive, negative classes, respectively.
By setting $\gamma^+ < \gamma^-$, we are able to concentrate the optimization on making correct classifications on positive classes.

Furthermore, to make additional efforts for reducing the effect of very easy negative samples, probability shifting is introduced:
\begin{equation}
    \label{eq:prob_shres}
     \hat{y}_l = \textrm{max}(\hat{y}_l - m, 0)
\end{equation}
where $m$ is a hyper-parameter that controls how many negative samples to drop.
When a class probability is below $m$, we drop any loss penalty for the class.

%% file: method.tex
\begin{figure*}[t]
    \centering
    \includegraphics[width=0.3\textwidth]{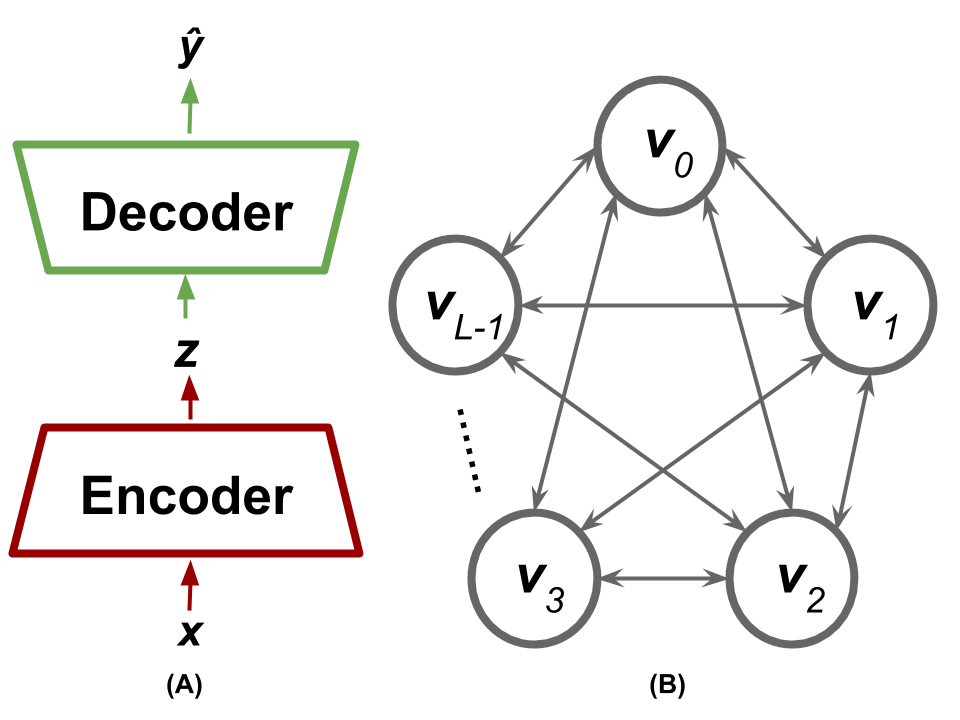}
    \includegraphics[width=0.3\textwidth]{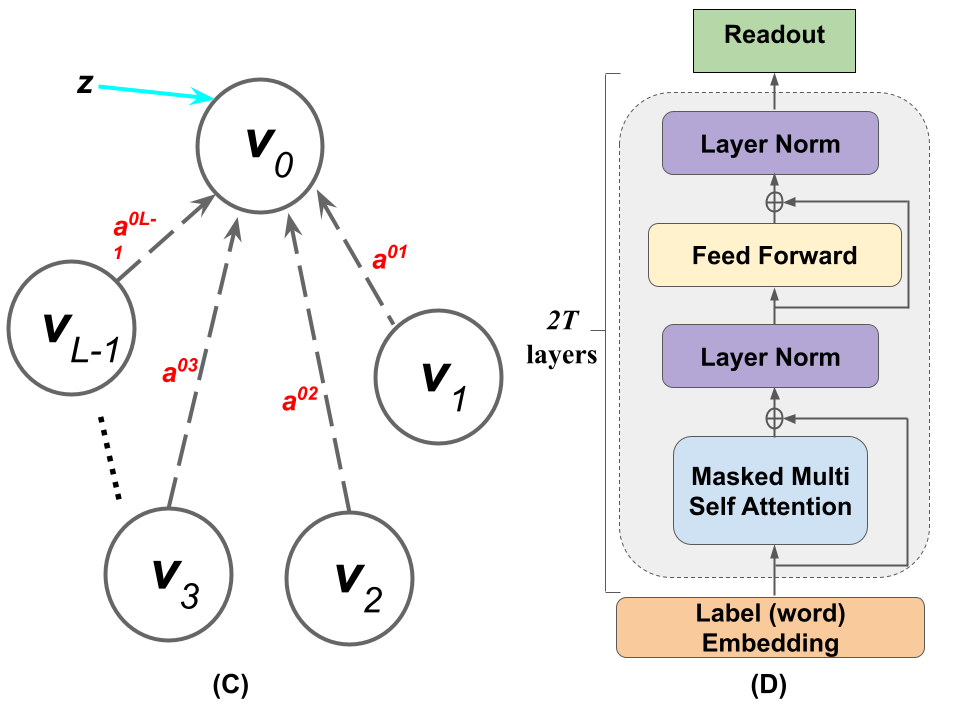}
    \caption{Model architecture of CbMLC. \textbf{(A)} Overall network architecture is in an encoder-to-decoder style. The encoder can be parameterized by any function depending on the application domain, and the decoder is a message passing network. \textbf{(B)} The underlying graph for the decoder. Each node corresponds to a label represented by an embedded vector, and there is an edge connecting every pair of nodes in both directions, meaning messages are passed both ways. 
    \textbf{(C)} An example of label node obtaining information from both $\bm{z}$ and other label nodes. Label 0 first receives $\bm{z}$ and collects messages from all other nodes with the attention weights $a^{01}$ through $a^{0L-1}$. \textbf{(D)} Decoder layers. Each of the $T$ layers computes a multi-head attention, followed by a feedforward operation. Layer normalization is used throughout to stabilize training. Finally, a readout layer is applied to produce the final label probability.} 
    \label{fig:framework}
\end{figure*}

\section{A Context-Based Multi-label Classifier}
We propose CbMLC, an end-to-end learning framework for multi-label classification, extending robust predictive performance on real-world datasets with noisy labels.
The key to its robustness is that CbMLC exploits rich contextual information in large-scale pretrained word embeddings by encouraging label embeddings similar to but not the same as word embeddings.
The context-based regularization provides the following benefits:
(1) over-parameterized neural networks can easily overfit noisy and less frequent labels, and incorporating word similarity information from pretrained word embeddings helps CbMLC to avoid spurious label correlations;
(2) when we are given a dataset with zero prior knowledge, the pretrained word embeddings guide learning in a promising direction;
and (3) regularizing based on the word embeddings alleviates the over-smoothing problem in GNNs.
In addition to the regularization, CbMLC also leverages recent advances in MLC such as asymmetric losses and employs self-attention in label graphs.

\label{sec:exp}
\subsection{Context-Based Regularization}
\label{sec:reg}

The key to context-based regularization is that we inject the word similarities learned by pre-trained word embeddings on label embeddings, which is done by adjusting the $\mathcal{L}2$ distance between the word embeddings and the label embeddings.
To begin, we have a word embedding function $g: \Delta \mapsto \mathbb R^P$, where $\Delta$ is a dictionary of word tokens, and $P$ is the dimensionality of the embedding space.
$g(\cdot)$ is often pretrained from large online corpora in an unsupervised fashion; thus, acquiring this contextual information does not require any additional supervision.
We denote a label embedding by $\bm{e}_l$ for the $l$th label.
To construct initial label embeddings, we define a mapping $h$ from word tokens of a label class $s_l$ to its embedding in $\mathbb R^P$.
If $s_l$ is a single word token $d$ and $s_l=d \in \Delta$, then $h(s_l) = g(d)$; however, if $s_l \notin \Delta$, then $h(s_l) \sim N(\mu, \sigma)^P$, where $\mu, \sigma$ are the mean and the variance of all the label tokens in $\Delta$. 
If $s_l$ is a label of multiple word tokens $\{d1, ..., d_K\}$, then $h(s_l) = \frac{1}{K}\sum_1^K h(d_k)$.
$\bm{e}_l$ is initially set as $h(s_l)$ and updated at every gradient step.
We define the context-based regularization as follows:
\begin{equation}
    \mathcal{L}_{CB} = \sum_{l=0}^{L-1} (\bm{e}_l - h(s_l))^2
\end{equation}
With this regularizer, CbMLC can learn label embeddings customized to each dataset while still maintaining some degree of global word similarities obtained from pre-trained word embeddings to prevent over-fitting.

\subsection{Model Architecture}
Figure~\ref{fig:framework} presents the model architecture of the CbMLC framework.
In Figure~\ref{fig:framework}A, the overall network flow follows an encoder-to-decoder structure: the encoder first maps input $x$ to a latent space, and the decoder takes the latent vector $\bm{z}$ to produce the final label probabilities.
The encoder can be parameterized by any function depending on the specific tasks, and the decoder is an attention-based MPNN on a label graph, where each node corresponds to a label.
From Figure~\ref{fig:framework}B, we can see that the label graph is by default a complete graph, where every pair of nodes is connected by an edge in both directions.
We can also remove edges if two labels are known to have no correlations.

In Figure~\ref{fig:framework}C, label node 0 first receives $\bm{z}$ and treats it as a single message to make an updates.
Then, label node 0 collects messages from its neighbors based on their attention weights to make the second update.
Figure~\ref{fig:framework}D demonstrates how to make one update.
Each label node has hidden states that are represented as embedded vectors $\{\bm{v}^t_0, \bm{v}^t_1, \dots, \bm{v}^t_{L-1}\}$, where $\bm{v}^t_l \in \mathbb R^P$ and $t$ is the time step, and $\bm{v}^0_l=\bm{e}_l$.
Self-attention is computed based on Equations~\ref{eq:alpha}-\ref{eq:mt}, and feedforward is computed using Equation~\ref{eq:u_attention}.
Specifically, we use multi-head self-attention~\cite{vaswani2017attention}, so that a node can focus on different parts simultaneously.
Each of the attention and feedforward sublayers is followed by layer normalization~\cite{ba2016layer} to reduce training instability.
Because at every $t$, $\bm{v}^t_l$ needs to update twice, there are $2T$ layers.
After the $2T$ layers, a readout layer predicts each label $\hat{y}_l$, where a readout function $R$ projects $\bm{v}_l^{2T}$ using a projection matrix $\textbf{W}^o \in \mathbb R^{d\times d}$.
The $l$th row of $\textbf{W}^o$ is denoted by $\textbf{W}^o_l$.
The resulting vector of size $L \times 1$ is then fed through an element-wise sigmoid function to produce the final probabilities of all labels:
\begin{equation}
   \label{eq:readout}
    \hat{y}_l = R({\bm{v}}^{2T}_l;{\textbf{W}^o}) = \textrm{sigmoid}({\textbf{W}^o_l} {\bm{v}}^{2T}_l).
\end{equation}
We note that $R(\cdot)$ is shared across all nodes; therefore, the prediction result is invariant to any node permutation.

\subsection{Loss Function}
The loss function $\mathcal{L}$ consists of two parts: the first part is a 0/1 loss that penalizes incorrect predictions, and the second part is the context-based (CB) regularizer described in Sec.~\ref{sec:reg}.
$\mathcal{L}$ is as follows:
\begin{equation}
    \mathcal{L} = \mathcal{L}_{\textrm{ASL}} + \lambda \mathcal{L}_{CB}
\end{equation}
$\mathcal{L}_{ASL}$ is the asymmetric cross entropy loss defined in Sec.~\ref{sec:background}.
Hyperparameter $\lambda$ controls to which degree we encourage labels to follow the word similarity structure from the pretrained word embbedings.
In a dataset with more training examples, we can set a small $\lambda$ as CbMLC has more examples to learn how labels interact with each other.
In summary, the parameters to be optimized are the model parameters together with the label embeddings.

%% file: experiment.tex
\begin{table*}[t]
\centering 
\resizebox{0.8\textwidth}{!}{%
\begin{tabular}{cc"c|c|c|c|c|c|c|c||c}
\textbf{Dataset}                    & \textbf{F1 score} & \textbf{MLKNN}  & \textbf{SLEEC}  & \textbf{Seq2Seq}         & \textbf{LaMP}  & \textbf{MPVAE}           & \textbf{ML-GCN}               & \textbf{resnet101-ce} & \textbf{resnet101-ASL}        & \textbf{CbMLC}           \\ \thickhline
\multirow{3}{*}{\textbf{bibtex}}    & \textbf{ebF1}     & 0.1826 & 0.4490  & 0.393           & 0.447 & 0.4534          & -                    & -            & -                    & {\ul \textit{\textbf{0.4566}}}\\
                           & \textbf{miF1}     & 0.1782 & 0.4074 & 0.384           & 0.473 & 0.4800            & -                    & -            & -                    & {\ul \textit{\textbf{0.4837}}}\\
                           & \textbf{maF1}     & 0.0727 & 0.2937 & 0.282           & 0.3760 & 0.3863          & -                    & -            & -                    & {\ul \textit{\textbf{0.4014}}}\\ \hline
\multirow{3}{*}{\textbf{reuters}}   & \textbf{ebF1}     & -      & -      & 0.8917          & 0.8950 & -               & -                    & -            & -                & {\ul \textit{\textbf{0.9064}}}      \\
                           & \textbf{miF1}     & -      & -      & 0.8545          & 0.8770 & -               & -                    & -            & - & {\ul \textit{\textbf{0.8893}}}      \\
                           & \textbf{maF1}     & -      & -      & 0.4567          & 0.5600  & -               & -                    & -            & - & {\ul \textit{\textbf{0.6024}}}      \\ \hline
\multirow{3}{*}{\textbf{delicious}} & \textbf{ebF1}     & 0.2590  & 0.3081 & 0.3200            & 0.3720 & {\ul \textit{\textbf{0.3732}}} & -                    & -            & -                    & 0.3558               \\
                           & \textbf{miF1}     & 0.2639 & 0.3333 & 0.3290           & 0.3860 & {\ul \textit{\textbf{0.3934}}} & -                    & -            & -                    & 0.3699               \\
                           & \textbf{maF1}     & 0.0526 & 0.1418 & 0.1660           & 0.196 & 0.1814          & -                    & -            & -                    & {\ul \textit{\textbf{0.1984}}}      \\ \hline
\multirow{3}{*}{\textbf{rcv1}}      & \textbf{ebF1}     & -      & -      & {\ul \textit{\textbf{0.8899}}} & 0.8870 & -               & -                    & -            & -                    & 0.8868               \\
                           & \textbf{miF1}     & -      & -      & {\ul \textit{\textbf{0.8797}}} & 0.8770 & -               & -                    & -            & -                    & 0.8735               \\
                           & \textbf{maF1}     & -      & -      & 0.7258          & 0.7400  & -               & -                    & -            & -                    & {\ul \textit{\textbf{0.7460}}}       \\ \hline
\multirow{3}{*}{\textbf{VOC}}       & \textbf{ebF1}     & -      & -      & -               & -     & -               & 0.9019               & 0.8903       & 0.8839               & {\ul \textit{\textbf{0.9067}}}\\
                           & \textbf{miF1}     & -      & -      & -               & -     & -               & 0.8904               & 0.8800         & 0.8743               & {\ul \textit{\textbf{0.8933}}} \\
                           & \textbf{maF1}     & -      & -      & -               & -     & -               & 0.8784               & 0.8657       & 0.8599               & {\ul \textit{\textbf{0.8822}}} \\ \hline
\end{tabular}%
}
\caption{Performance comparison (based on example- and label-based F1 scores) of CbMLC to state-of-the-art models on five datasets with clean labels. Higher scores are better. Highlighted numbers are the best of in each row.}
\label{tab:main_exps}
\end{table*}

\section{Experiments}

\begin{figure*}[h]
    \centering
    \includegraphics[width=0.8\textwidth]{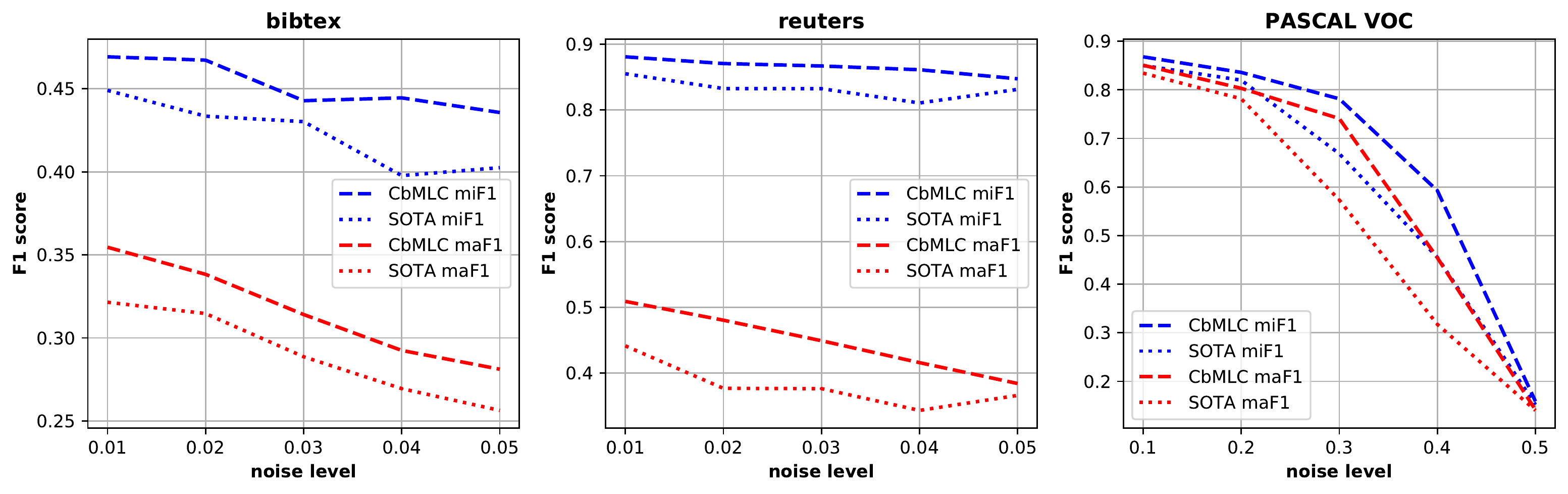}
    \includegraphics[width=0.8\textwidth]{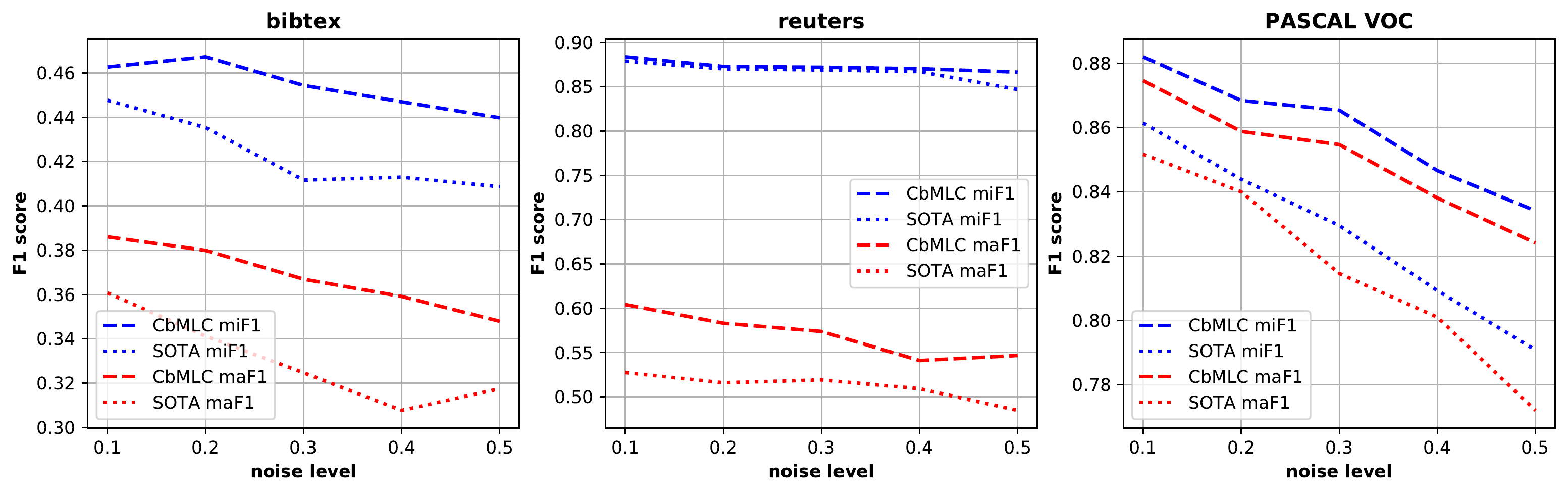}
    \caption{Performance comparison (based on label-based F1 scores) of CbMLC to SOTA MLC models on three datasets with uniform noise across all labels \textbf{(top)} and with uniform noise across positive labels \textbf{(bottom)}. The blue lines are miF1 scores, and the red lines are the maF1 scores. The dashed lines represent CbMLC; the dotted lines represent the best SOTA methods for the corresponding datasets.}
    \label{fig:exps_uniformpos}
\end{figure*}

We perform a thorough comparison of CbMLC to the state of the art (SOTA) on a variety of datasets ranging from NLP to CV under both clean and noisy settings.
We also present an ablation study to demonstrate the effectiveness of the context-based regularization.

\paragraph{Datasets and Evaluation Metrics.}
We run experiments on the following datasets available for download online~\footnote{http://mulan.sourceforge.net/datasets-mlc.html}~\footnote{http://host.robots.ox.ac.uk/pascal/VOC/}: \emph{reuters-21578} and \emph{rcv1-v2}~\cite{lewis2004rcv1}, natural language sequences with predefined categories based on their content, \emph{bibtex}~\cite{katakis2008multilabel} and \emph{delicious}~\cite{pmlr-v28-bi13}, collections of unordered text objects associated with tags, and \emph{Pascal VOC}~\cite{Everingham15}, RGB images with multiple objects simultaneously present.
For \emph{reuters-21578}, \emph{rcv1-v2}, \emph{bibtex}, and \emph{delicious}, we split them into a training set, a validation set, and a test set in an identical way as that in \citet{bai2020disentangled} and \citet{lanchantin2019neural}.
For \emph{Pascal VOC} (for brevity, we later refer to it as \emph{VOC}), we follow the exact training settings used in \citet{chen2019multi} and \citet{ben2020asymmetric}.

Following conventional MLC settings, we report example-based F1 (ebF1) scores, label-based micro-averaged F1 scores, and label-based macro-averaged F1 scores.
When there are imbalanced classes, high miF1 scores indicate better performance on frequent labels, and high maF1 scores indicate strong results on less frequent labels.

\paragraph{Baseline Comparisons.} We compare CbMLC with domain specific SOTA MLC methods: Seq2Seq~\cite{nam2017maximizing} and LaMP~\cite{lanchantin2019neural} are two SOTA approaches for multi-label text classification, and ML-GCN~\cite{chen2019multi} is the SOTA in image recognition.
MLKNN~\cite{zhang2007ml}, SLEEC~\cite{NIPS2015_5969}, and MPVAE~\cite{bai2020disentangled} are general MLC techniques, but they cannot handle image or sequence inputs.
For CV datasets, we also compare to the resnet101~\cite{He_2016_CVPR}, with either BCE or ASL as the loss function.
Among all baseline methods, \emph{MPVAE claims to be a noise-robust model}.
We follow the same training settings and tune hyper-parameters to achieve the best performance for every baseline method.
We note that, we do not directly compare to single-label methods for dealing with noise because 1) many such works assume a softmax output layer~\cite{zhang2018generalized,Lyu2020Curriculum,jindal2019effective} or access to a small set of clean data~\cite{zhang2020distilling}, 
and 2) CbMLC is general in that the decoder and the context-based regularization can be combined with any feature extractor, so CbMLC can be incorporated with other noise learning methods such as meta-learning approaches~\cite{garcia2016noise}.
However, we do modify one SOTA single-label noise method using loss correction and include the comparison results in supplementary material.

\paragraph{Training and Implementation Details, etc.}
We run experiments on NVIDIA Tesla V100 GPUs, each with 16GB memory.
We train CbMLC for up to 50 epochs with the Adam~\cite{DBLP:journals/corr/KingmaB14} optimizer and use a weight decay of $1e^{-5}$.
The number of decoder layers is 4.
For NLP datasets, we adopt a transformer-style encoder used in \citet{lanchantin2019neural} and set the learning rate to be $2^{-4}$.
For CV datasets, we use resnet101~\cite{He_2016_CVPR} as the encoder, and the learning rate is $1e^{-5}$.
CbMLC only requires minimal tuning; we only search $\lambda$ from $\{0.001, 0.01, 0.1, 1\}$ and $\gamma^{+}$ as well as $\gamma^{-}$ from $\{1, 2, 4\}$.
$m$ is $0.05$ or $0.1$.

\subsection{Experiments on Clean Datasets}
Table~\ref{tab:main_exps} presents the performance comparison of CbMLC to other SOTA MLC methods on the five datasets.
On all five datasets, CbMLC produces superior or at least comparable performance to that of previous works.
Furthermore, we have the highest maF1 score for every dataset; that is, our method tends to learn better for the classes with less positive labels.
CbMLC produces relatively smaller improvements in miF1 scores; this could be because there are already many samples with frequent labels, so it is possible to learn how the labels interact from label co-occurrences alone.
\subsection{Experiments on Datasets with Injected Noise}
We compare CbMLC to other SOTA MLC methods with injected noise discussed in Section~\ref{sec:noise}.
For type 1 noise, we test with $\{1\%,2\%,3\%,4\%,5\%\}$ uniform noise for the NLP datasets and $\{10\%,20\%,30\%,40\%,50\%\}$ uniform noise for the CV datasets across both positive and negative labels (since there are far fewer label classes for the CV datasets, adding as little noise as we add to the NLP datasets leads to very minor drop in performance).
For type 2 noise, we test with $\{10\%,20\%,30\%,40\%,50\%\}$ uniform noise on all the datasets across all positive labels.
We also test on type 3 noise, where each training sample only has one positive label that is randomly chosen.
Due to the page limit, we select one dataset from each task to present the results for type 1 and type 2 noise.
We include comprehensive results for all the datasets in the supplementary material.
To produce a noisy setting close to the real world, we also test the methods with combined noise: for each instance, there is 1/3 chance for it to be corrupted by one type of noise; for type 1 noise, 0\% to 10\% positive and negative labels are modified, and for type 2 noise, 0\% to 50\% positive negative labels are modified.

We present the results for type 1 noise and type 2 noise in Figure~\ref{fig:exps_uniformpos}.
The dashed lines show the performance of CbMLC, and the dotted lines show the performance of the best performing SOTA MLC method on the corresponding dataset.
The blue lines represent the miF1 scores, and the red lines are the maF1 scores.
In most cases, we produce a substantial improvement.
On \emph{reuters} and \emph{bibtex}, light noise (e.g., 1\%, 2\% type 1 noise and 10\%, 20\% type 2 noise) does not lead to much drop in the classification results; they are even comparable to running other SOTA methods on the clean datasets.
We also highlight how our method performs on \emph{Pascal VOC}.
When there is 30\% type 1 noise, we improve the maF1 score by the best SOTA method by nearly 30\%, and when there is 40\% type 1 noise, we improve the miF1 score by the best SOTA method by 30\% as well.

We summarize the results for type 3 noise in Table~\ref{tab:exps_onepos} and the results for the combined noise in Table~\ref{tab:exps_combined}.
We outperform the best SOTA method on every dataset (except for the miF1 score on \emph{rcv1} with type 3 noise).
When the noise becomes more complex, the performance gaps between the best SOTA method and CbMLC further increase.
Therefore, context-based regularization is a simple yet very effective trick to prevent overfit label noise.
To summarize, CbMLC produce relativelly small improvements on the clean datasets; however, when there are noisy labels, CbMLC is more robust than other SOTA methods.

\begin{table}[t]
\centering
\resizebox{0.45\textwidth}{!}{%
\begin{tabular}{cc"c|c|c|c|c}
\multicolumn{1}{l}{}           & \textbf{F1 score} & \textbf{bibtex} & \textbf{reuters} & \textbf{delicious} & \textbf{rcv1} & \textbf{VOC} \\ \thickhline
\multirow{2}{*}{\textbf{SOTA}} & \textbf{miF1}                 & 0.3572                         & 0.8165                         & 0.2737                         & {\ul \textit{\textbf{0.8483}}}                         & 0.8350\\
                               & \textbf{maF1}                 & 0.2387                         & 0.4000                            & 0.0363                         & 0.6770                          & 0.8302\\ \hline
\multirow{2}{*}{\textbf{CbMLC}}   & \textbf{miF1}                 & {\ul \textit{\textbf{0.4210}}}  & {\ul \textit{\textbf{0.8581}}} & {\ul \textit{\textbf{0.2805}}} & 0.8448 & {\ul \textit{\textbf{0.8447}}}\\
                               & \textbf{maF1}                 & {\ul \textit{\textbf{0.3049}}} & {\ul \textit{\textbf{0.4758}}} & {\ul \textit{\textbf{0.0636}}} & {\ul \textit{\textbf{0.7007}}} & {\ul \textit{\textbf{0.8442}}}                             
\end{tabular}%
}
\caption{Performance comparison (based on label-based F1 scores) of CbMLC to SOTA MLC models on the five datasets with injected noise where each training sample only has one positive label.}
\label{tab:exps_onepos}
\end{table}
\begin{table}[t]
\centering
\resizebox{0.45\textwidth}{!}{%
\begin{tabular}{cc"c|c|c|c|c}
\multicolumn{1}{l}{}            & \textbf{F1 score} & \textbf{bibtex}          & \textbf{reuters}         & \textbf{delicious}       & \textbf{rcv1}       & \textbf{VOC}      \\ \thickhline
\multirow{2}{*}{\textbf{SOTA}}  & \textbf{miF1}        & 0.3238                   & 0.7484                   & 0.1973                   & 0.8579                   & 0.7138                   \\
                                & \textbf{maF1}        & 0.1888                   & 0.2768                   & 0.0682                   & 0.6646                   & 0.6281                   \\
\multirow{2}{*}{\ul\textbf{CbMLC}} & \textbf{miF1}        & {\ul\textit{\textbf{0.3282}}} & {\ul\textit{\textbf{0.8200}}}   & {\ul\textit{\textbf{0.3290}}}  & {\ul\textit{\textbf{0.8580}}} & {\ul\textit{\textbf{0.7621}}} \\
                                & \textbf{maF1}        & {\ul\textit{\textbf{0.1993}}} & {\ul\textit{\textbf{0.3325}}} & {\ul\textit{\textbf{0.0811}}} & {\ul\textit{\textbf{0.6735}}} & {\ul\textit{\textbf{0.7281}}}
\end{tabular}%
}
\caption{Performance comparison (based on label-based F1 scores) of CbMLC to SOTA MLC models on the five datasets injected with combined noise types.}
\label{tab:exps_combined}
\end{table}

\begin{figure}[t]
    \centering
    \includegraphics[width=0.35\textwidth]{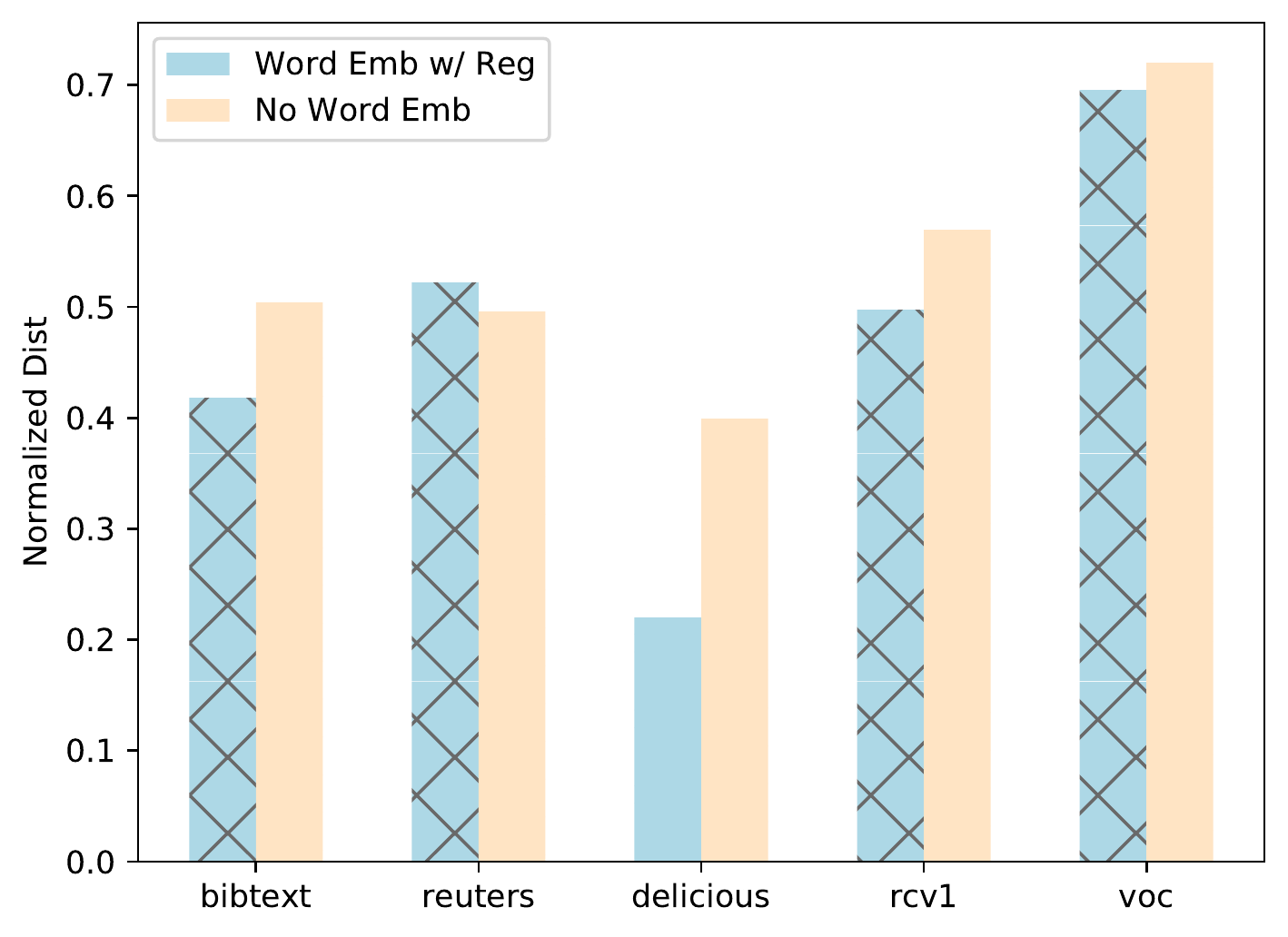}
    \caption{Embedding distances between label pairs averaged over top 100 most frequently co-occurring label pairs for each dataset. A smaller distance for a label pair indicates the two labels have stronger correlation. \textit{Using word embeddings with the context-based regularizer for label embeddings produces smaller distances for label pairs with frequent co-occurrences than not using word embeddings, suggesting CbMLC learns better label semantics.}}
    \label{fig:emb_dist}
\end{figure}

\begin{table}[h]
\centering
\resizebox{0.45\textwidth}{!}{%
\begin{tabular}{cc"c|c|c|c|c}
\multicolumn{1}{l}{}                                                                    & \textbf{F1 score} & \textbf{bibtext}               & \textbf{reuters}               & \textbf{delicious}             & \textbf{rcv1}                  & \textbf{VOC} \\ \thickhline
\multirow{3}{*}{\textbf{\begin{tabular}[c]{@{}c@{}}Fixed\\ WordEmb\end{tabular}}}       & \textbf{ebF1}  & 0.4484                         & 0.9045                         & {\ul \textit{\textbf{0.3634}}} & 0.8867                         & 0.9023\\
                                                                                        & \textbf{miF1}  & 0.4814                         & 0.8830                          & {\ul \textit{\textbf{0.3749}}} & 0.8734                         & 0.8901\\
                                                                                        & \textbf{maF1}                          & 0.3834                         & 0.5530                          & 0.1949                         & 0.7320                          & 0.8797\\ \hline
\multirow{3}{*}{\textbf{\begin{tabular}[c]{@{}c@{}}Regularized\\ WordEmb\end{tabular}}} & \textbf{ebF1}                          & {\ul \textit{\textbf{0.4538}}} & {\ul \textit{\textbf{0.9064}}} & 0.3558                         & {\ul \textit{\textbf{0.8868}}} & {\ul \textit{\textbf{0.9049}}}\\
                                                                                        & \textbf{miF1}                          & {\ul \textit{\textbf{0.4893}}} & {\ul \textit{\textbf{0.8893}}} & 0.3699                         & {\ul \textit{\textbf{0.8735}}} & {\ul \textit{\textbf{0.8919}}}\\
                                                                                        & \textbf{maF1}                          & {\ul \textit{\textbf{0.3940}}}  & {\ul \textit{\textbf{0.6024}}} & {\ul \textit{\textbf{0.1984}}} & {\ul \textit{\textbf{0.7460}}}  & {\ul \textit{\textbf{0.8822}}}
\end{tabular}%
}
\caption{Comparison on whether the context-based regularization yields an improvement over having fixed word embeddings. Highlighted numbers are the best in each row.}
\label{tab:emb_effect}
\end{table}
\subsection{Ablation Study}
\paragraph{Regularized vs. Fixed Word Embeddings.}
To verify context-based regularization provides a benefit, we first show a performance comparison of applying regularized word embeddings to applying fixed word embeddings as label embeddings in Table~\ref{tab:emb_effect}.
Except for \emph{delicious}, all other datasets favor using context-based regularization.

\paragraph{Measuring Label Correlations.}
Additionally, we measure the Euclidean distance between pairs of label embeddings.
In particular, we sort label pairs based on the number of times they co-occur in the training set and compute the average distance for the 100 pairs with most co-occurrences for the label embeddings with random initialization and for the label embeddings with context-based regularization.
We observe that training with randomly initialized label embeddings often suffers from a more severe over-smoothing problem; in order to reduce this effect, we normalize the distance before comparing Euclidean distances across different embedding spaces. 
For a highly correlated label pair, having a closer distance in the embedding space suggests the model successfully captures their correlation.
We present the results in Figure~\ref{fig:emb_dist}.
Out of the five datasets, we learn more accurate label similarities on four of them.

%% file: related.tex
\section{Related Work}
Existing MLC approaches that are related to our work incorporate label correlations, use regularization or an unconventional loss, or consider MLC under a noisy label setting.

\paragraph{Learning Label Dependencies.}
Our technique is within the line of work where conditional label dependencies are considered when making label predictions, as opposed to the methods that decompose a MLC problem into a set of independent binary classification problems~\cite{boutell2004learning,zhang2007ml}.


Probabilistic chain classifiers (PCCs) stack binary classifiers sequentially and produce one label at a time conditioned on all previous outputs~\cite{cheng2010bayes,read2011classifier}.
Followup works extend PCCs to recurrent neural networks (RNNs)~\cite{wang2016cnn,nam2017maximizing}.
There are two major drawbacks of these proposed models.
They are not parallelizable due to the auto-regressive nature, so either training or performing inference is time-cosuming.
Additionally, MLC labels are unordered sets, but sequential prediction is highly order-dependent; therefore, ordering labels differently could significantly impact performance.

\citeauthor{chen2017deep,chen2018end,bai2020disentangled} develop deep multi-variate probit models (DMVPs) to tackle MLC problems.
DMVPs first compute a probability of each label being present and use a label covariance matrix to make adjustments.
However, because of covariance matrices, DMVPs can only model pairwise label correlations, and real-world label interactions often happen beyond second order~\cite{wang2020multi}.

Building label graphs on GNNs overcome the prior obstacles: their graphical structures enables to learn label dependencies in an order-invariant manner, and they have the potential to capture high-order label dependencies by stacking multiple layers together.
\citeauthor{chen2019multi,10.1145/3394486.3403368,DBLP:conf/aaai/WangHLLZMW20,wang2018joint} represent each node of the graph as a word embedding of the corresponding label.
This may be problematic: word embeddings are fixed for all datasets, but the pretrained word embeddings are unlikely to fully capture the relations between words for every dataset.
The learning capacities are thus degraded.
\citeauthor{lanchantin2019neural,wang2020multi,DBLP:conf/aaai/YouGCLBW20} model label interactions using attention-based message passing neural networks (MPNNs)~\cite{gilmer2017neural}, where they learn label representations from scratch.
 However, this may easily overfit when there are noisy labels. 


\paragraph{Losses and Regularization Techniques.}
In addition to BCE, other losses have been proposed to train multi-label classifiers.
\citet{gong2014deep} perform extensive experiments on evaluating softmax~\cite{guillaumin2009tagprop}, pairwise ranking~\cite{joachims2002optimizing}, and weighted approximate ranking~\cite{weston2011wsabie} for MLC.
\cite{lin2017focal} present focal loss to weight positive and negative labels differently.
Regularization techniques are an important research direction to prevent overfitting on training sets.
\citet{zhang2017learning,krichene2018efficient,guo2019breaking} maximize the distances between all pairs of label embeddings to account for sparse labels in extreme MLC.

\paragraph{MLC with Noisy Labels}
To handle noisy labels in MLC, \citet{garcia2016noise} propose to adjust incorrectly estimated label transition probabilities by quality embedding.
Partial multi-label learning (PMLL) setting is also related~\cite{xie2018partial}, where each example is annotated with a candidate label set that contains relevant labels and noisy labels.
To identify the noisy labels, \cite{zhang2020partial} recover true labels by introducing labeling confidence, \cite{sun2019partial} exploit a low-rank and sparse decomposition scheme, and \cite{Xie_Huang_2020} optimize the multi-label classifier and noisy label identifier under a unified framework.

%% file: conclusion.tex
\section{Conclusion}
While label noise is a common problem in MLC, few works have evaluated MLC algorithms with noisy labels.
We identify three common noise scenarios and create a noise setting to perform evaluations with combined noise types.
As the MLC methods that learn label interactions can easily be affected by noisy data, we present CbMLC, which employs an asymmetric loss function with context-based regularization.
Our experiments show substantial improvements over other SOTA MLC techniques on the noisy datasets.

%% file: icml2021.bbl
\begin{thebibliography}{70}
\providecommand{\natexlab}[1]{#1}
\providecommand{\url}[1]{\texttt{#1}}
\expandafter\ifx\csname urlstyle\endcsname\relax
  \providecommand{\doi}[1]{doi: #1}\else
  \providecommand{\doi}{doi: \begingroup \urlstyle{rm}\Url}\fi

\bibitem[Ba et~al.(2016)Ba, Kiros, and Hinton]{ba2016layer}
Ba, J.~L., Kiros, J.~R., and Hinton, G.~E.
\newblock Layer normalization.
\newblock \emph{arXiv preprint arXiv:1607.06450}, 2016.

\bibitem[Bai et~al.(2020)Bai, Kong, and Gomes]{bai2020disentangled}
Bai, J., Kong, S., and Gomes, C.
\newblock Disentangled variational autoencoder based multi-label classification
  with covariance-aware multivariate probit model.
\newblock In \emph{Proceedings of the Twenty-Ninth International Joint
  Conference on Artificial Intelligence, {IJCAI-20}}, pp.\  4313--4321, 7 2020.
\newblock \doi{10.24963/ijcai.2020/595}.
\newblock Special track on AI for CompSust and Human well-being.

\bibitem[Beigman~Klebanov \& Beigman(2009)Beigman~Klebanov and
  Beigman]{beigman2009annotator}
Beigman~Klebanov, B. and Beigman, E.
\newblock From annotator agreement to noise models.
\newblock \emph{Computational Linguistics}, 35\penalty0 (4):\penalty0 495--503,
  2009.

\bibitem[Ben-Baruch et~al.(2020)Ben-Baruch, Ridnik, Zamir, Noy, Friedman,
  Protter, and Zelnik-Manor]{ben2020asymmetric}
Ben-Baruch, E., Ridnik, T., Zamir, N., Noy, A., Friedman, I., Protter, M., and
  Zelnik-Manor, L.
\newblock Asymmetric loss for multi-label classification.
\newblock \emph{arXiv preprint arXiv:2009.14119}, 2020.

\bibitem[Bhatia et~al.(2015)Bhatia, Jain, Kar, Varma, and Jain]{NIPS2015_5969}
Bhatia, K., Jain, H., Kar, P., Varma, M., and Jain, P.
\newblock Sparse local embeddings for extreme multi-label classification.
\newblock In Cortes, C., Lawrence, N.~D., Lee, D.~D., Sugiyama, M., and
  Garnett, R. (eds.), \emph{Advances in Neural Information Processing Systems
  28}, pp.\  730--738. Curran Associates, Inc., 2015.

\bibitem[Bi \& Kwok(2013)Bi and Kwok]{pmlr-v28-bi13}
Bi, W. and Kwok, J.
\newblock Efficient multi-label classification with many labels.
\newblock In Dasgupta, S. and McAllester, D. (eds.), \emph{Proceedings of the
  30th International Conference on Machine Learning}, volume~28 of
  \emph{Proceedings of Machine Learning Research}, pp.\  405--413, Atlanta,
  Georgia, USA, 17--19 Jun 2013. PMLR.
\newblock URL \url{http://proceedings.mlr.press/v28/bi13.html}.

\bibitem[Boutell et~al.(2004)Boutell, Luo, Shen, and
  Brown]{boutell2004learning}
Boutell, M.~R., Luo, J., Shen, X., and Brown, C.~M.
\newblock Learning multi-label scene classification.
\newblock \emph{Pattern recognition}, 37\penalty0 (9):\penalty0 1757--1771,
  2004.

\bibitem[Chang et~al.(2020)Chang, Yu, Zhong, Yang, and
  Dhillon]{10.1145/3394486.3403368}
Chang, W.-C., Yu, H.-F., Zhong, K., Yang, Y., and Dhillon, I.~S.
\newblock \emph{Taming Pretrained Transformers for Extreme Multi-Label Text
  Classification}, pp.\  3163–3171.
\newblock Association for Computing Machinery, New York, NY, USA, 2020.
\newblock ISBN 9781450379984.

\bibitem[Chen et~al.(2017)Chen, Xue, Fink, Chen, and Gomes]{chen2017deep}
Chen, D., Xue, Y., Fink, D., Chen, S., and Gomes, C.~P.
\newblock Deep multi-species embedding.
\newblock In \emph{Proceedings of the 26th International Joint Conference on
  Artificial Intelligence}, pp.\  3639--3646, 2017.

\bibitem[Chen et~al.(2018)Chen, Xue, and Gomes]{chen2018end}
Chen, D., Xue, Y., and Gomes, C.
\newblock End-to-end learning for the deep multivariate probit model.
\newblock In \emph{International Conference on Machine Learning}, pp.\
  932--941, 2018.

\bibitem[Chen et~al.(2019)Chen, Wei, Wang, and Guo]{chen2019multi}
Chen, Z.-M., Wei, X.-S., Wang, P., and Guo, Y.
\newblock Multi-label image recognition with graph convolutional networks.
\newblock In \emph{Proceedings of the IEEE Conference on Computer Vision and
  Pattern Recognition}, pp.\  5177--5186, 2019.

\bibitem[Cheng et~al.(2010)Cheng, H{\"u}llermeier, and
  Dembczynski]{cheng2010bayes}
Cheng, W., H{\"u}llermeier, E., and Dembczynski, K.~J.
\newblock Bayes optimal multilabel classification via probabilistic classifier
  chains.
\newblock In \emph{Proceedings of the 27th international conference on machine
  learning (ICML-10)}, pp.\  279--286, 2010.

\bibitem[Chua et~al.(2009)Chua, Tang, Hong, Li, Luo, and
  Zheng]{10.1145/1646396.1646452}
Chua, T.-S., Tang, J., Hong, R., Li, H., Luo, Z., and Zheng, Y.
\newblock Nus-wide: A real-world web image database from national university of
  singapore.
\newblock In \emph{Proceedings of the ACM International Conference on Image and
  Video Retrieval}, CIVR '09, New York, NY, USA, 2009. Association for
  Computing Machinery.
\newblock ISBN 9781605584805.
\newblock \doi{10.1145/1646396.1646452}.
\newblock URL \url{https://doi.org/10.1145/1646396.1646452}.

\bibitem[Denoeux(2000)]{denoeux2000neural}
Denoeux, T.
\newblock A neural network classifier based on dempster-shafer theory.
\newblock \emph{IEEE Transactions on Systems, Man, and Cybernetics-Part A:
  Systems and Humans}, 30\penalty0 (2):\penalty0 131--150, 2000.

\bibitem[Everingham et~al.(2015)Everingham, Eslami, Van~Gool, Williams, Winn,
  and Zisserman]{Everingham15}
Everingham, M., Eslami, S. M.~A., Van~Gool, L., Williams, C. K.~I., Winn, J.,
  and Zisserman, A.
\newblock The pascal visual object classes challenge: A retrospective.
\newblock \emph{International Journal of Computer Vision}, 111\penalty0
  (1):\penalty0 98--136, January 2015.

\bibitem[Fr{\'e}nay \& Verleysen(2013)Fr{\'e}nay and
  Verleysen]{frenay2013classification}
Fr{\'e}nay, B. and Verleysen, M.
\newblock Classification in the presence of label noise: a survey.
\newblock \emph{IEEE transactions on neural networks and learning systems},
  25\penalty0 (5):\penalty0 845--869, 2013.

\bibitem[Garcia et~al.(2016)Garcia, de~Carvalho, and Lorena]{garcia2016noise}
Garcia, L.~P., de~Carvalho, A.~C., and Lorena, A.~C.
\newblock Noise detection in the meta-learning level.
\newblock \emph{Neurocomputing}, 176:\penalty0 14--25, 2016.

\bibitem[Gilmer et~al.(2017)Gilmer, Schoenholz, Riley, Vinyals, and
  Dahl]{gilmer2017neural}
Gilmer, J., Schoenholz, S.~S., Riley, P.~F., Vinyals, O., and Dahl, G.~E.
\newblock Neural message passing for quantum chemistry.
\newblock \emph{arXiv preprint arXiv:1704.01212}, 2017.

\bibitem[Gong et~al.(2014)Gong, Jia, Toshev, Leung, and Ioffe]{gong2014deep}
Gong, Y., Jia, Y., Toshev, A., Leung, T., and Ioffe, S.
\newblock Deep convolutional ranking for multilabel image annotation.
\newblock In \emph{International Conference on Learning Representations}, 2014.

\bibitem[Guillaumin et~al.()Guillaumin, Mensink, Verbeek, and
  Schmid]{guillaumin2009tagprop}
Guillaumin, M., Mensink, T., Verbeek, J., and Schmid, C.
\newblock Tagprop: Discriminative metric learning in nearest neighbor models
  for image auto-annotation.
\newblock In \emph{2009 IEEE 12th international conference on computer vision},
  pp.\  309--316. IEEE.

\bibitem[Guo et~al.(2019)Guo, Mousavi, Wu, Holtmann-Rice, Kale, Reddi, and
  Kumar]{guo2019breaking}
Guo, C., Mousavi, A., Wu, X., Holtmann-Rice, D.~N., Kale, S., Reddi, S., and
  Kumar, S.
\newblock Breaking the glass ceiling for embedding-based classifiers for large
  output spaces.
\newblock In Wallach, H., Larochelle, H., Beygelzimer, A., d\textquotesingle
  Alch\'{e}-Buc, F., Fox, E., and Garnett, R. (eds.), \emph{Advances in Neural
  Information Processing Systems}, volume~32, pp.\  4943--4953. Curran
  Associates, Inc., 2019.

\bibitem[He et~al.(2016)He, Zhang, Ren, and Sun]{He_2016_CVPR}
He, K., Zhang, X., Ren, S., and Sun, J.
\newblock Deep residual learning for image recognition.
\newblock In \emph{Proceedings of the IEEE Conference on Computer Vision and
  Pattern Recognition (CVPR)}, June 2016.

\bibitem[Ipeirotis et~al.(2010)Ipeirotis, Provost, and
  Wang]{ipeirotis2010quality}
Ipeirotis, P.~G., Provost, F., and Wang, J.
\newblock Quality management on amazon mechanical turk.
\newblock In \emph{Proceedings of the ACM SIGKDD workshop on human
  computation}, pp.\  64--67, 2010.

\bibitem[Jindal et~al.(2019)Jindal, Pressel, Lester, and
  Nokleby]{jindal2019effective}
Jindal, I., Pressel, D., Lester, B., and Nokleby, M.
\newblock An effective label noise model for dnn text classification.
\newblock In \emph{Proceedings of the 2019 Conference of the North American
  Chapter of the Association for Computational Linguistics: Human Language
  Technologies, Volume 1 (Long and Short Papers)}, pp.\  3246--3256, 2019.

\bibitem[Joachims(2002)]{joachims2002optimizing}
Joachims, T.
\newblock Optimizing search engines using clickthrough data.
\newblock In \emph{Proceedings of the eighth ACM SIGKDD international
  conference on Knowledge discovery and data mining}, pp.\  133--142, 2002.

\bibitem[Joulin et~al.(2016)Joulin, Grave, Bojanowski, Douze, J{\'e}gou, and
  Mikolov]{joulin2016fasttext}
Joulin, A., Grave, E., Bojanowski, P., Douze, M., J{\'e}gou, H., and Mikolov,
  T.
\newblock Fasttext. zip: Compressing text classification models.
\newblock \emph{arXiv preprint arXiv:1612.03651}, 2016.

\bibitem[Katakis et~al.(2008)Katakis, Tsoumakas, and
  Vlahavas]{katakis2008multilabel}
Katakis, I., Tsoumakas, G., and Vlahavas, I.
\newblock Multilabel text classification for automated tag suggestion.
\newblock \emph{ECML PKDD Discovery Challenge 2008}, pp.\ ~75, 2008.

\bibitem[Kingma \& Ba(2015)Kingma and Ba]{DBLP:journals/corr/KingmaB14}
Kingma, D.~P. and Ba, J.
\newblock Adam: A method for stochastic optimization.
\newblock In \emph{ICLR (Poster)}, 2015.

\bibitem[Krasin et~al.(2017)Krasin, Duerig, Alldrin, Ferrari, Abu-El-Haija,
  Kuznetsova, Rom, Uijlings, Popov, Veit, Belongie, Gomes, Gupta, Sun, Chechik,
  Cai, Feng, Narayanan, and Murphy]{openimages}
Krasin, I., Duerig, T., Alldrin, N., Ferrari, V., Abu-El-Haija, S., Kuznetsova,
  A., Rom, H., Uijlings, J., Popov, S., Veit, A., Belongie, S., Gomes, V.,
  Gupta, A., Sun, C., Chechik, G., Cai, D., Feng, Z., Narayanan, D., and
  Murphy, K.
\newblock Openimages: A public dataset for large-scale multi-label and
  multi-class image classification.
\newblock \emph{Dataset available from
  https://storage.googleapis.com/openimages/web/index.html}, 2017.

\bibitem[Krichene et~al.(2019)Krichene, Mayoraz, Rendle, Zhang, Yi, Hong, Chi,
  and Anderson]{krichene2018efficient}
Krichene, W., Mayoraz, N., Rendle, S., Zhang, L., Yi, X., Hong, L., Chi, E.,
  and Anderson, J.
\newblock Efficient training on very large corpora via gramian estimation.
\newblock In \emph{International Conference on Learning Representations}, 2019.

\bibitem[Lanchantin et~al.(2019)Lanchantin, Sekhon, and
  Qi]{lanchantin2019neural}
Lanchantin, J., Sekhon, A., and Qi, Y.
\newblock Neural message passing for multi-label classification.
\newblock In \emph{Joint European Conference on Machine Learning and Knowledge
  Discovery in Databases}, pp.\  138--163. Springer, 2019.

\bibitem[Lewis et~al.(2004)Lewis, Yang, Rose, and Li]{lewis2004rcv1}
Lewis, D.~D., Yang, Y., Rose, T.~G., and Li, F.
\newblock Rcv1: A new benchmark collection for text categorization research.
\newblock \emph{Journal of machine learning research}, 5\penalty0
  (Apr):\penalty0 361--397, 2004.

\bibitem[Li et~al.(2021)Li, Xiong, and Hoi]{li2021towards}
Li, J., Xiong, C., and Hoi, S.
\newblock Towards noise-resistant object detection with noisy annotations,
  2021.
\newblock URL \url{https://openreview.net/forum?id=TlPHO_duLv}.

\bibitem[Li et~al.(2017)Li, Wang, Li, Agustsson, Berent, Gupta, Sukthankar, and
  Van~Gool]{li2017webvision}
Li, W., Wang, L., Li, W., Agustsson, E., Berent, J., Gupta, A., Sukthankar, R.,
  and Van~Gool, L.
\newblock Webvision challenge: visual learning and understanding with web data.
\newblock \emph{arXiv preprint arXiv:1705.05640}, 2017.

\bibitem[Lin et~al.(2017)Lin, Goyal, Girshick, He, and
  Doll{\'a}r]{lin2017focal}
Lin, T.-Y., Goyal, P., Girshick, R., He, K., and Doll{\'a}r, P.
\newblock Focal loss for dense object detection.
\newblock In \emph{Proceedings of the IEEE international conference on computer
  vision}, pp.\  2980--2988, 2017.

\bibitem[Liu et~al.(2017)Liu, Chang, Wu, and Yang]{liu2017deep}
Liu, J., Chang, W.-C., Wu, Y., and Yang, Y.
\newblock Deep learning for extreme multi-label text classification.
\newblock In \emph{Proceedings of the 40th International ACM SIGIR Conference
  on Research and Development in Information Retrieval}, pp.\  115--124, 2017.

\bibitem[Lyu \& Tsang(2020)Lyu and Tsang]{Lyu2020Curriculum}
Lyu, Y. and Tsang, I.~W.
\newblock Curriculum loss: Robust learning and generalization against label
  corruption.
\newblock In \emph{International Conference on Learning Representations}, 2020.
\newblock URL \url{https://openreview.net/forum?id=rkgt0REKwS}.

\bibitem[McNicol(1972)]{mcnicol1972statistical}
McNicol, D.
\newblock What are statistical decisions.
\newblock In \emph{A Primer of Signal Detection Theory}, pp.\  1--17. Allen,
  1972.

\bibitem[Mikolov et~al.(2013)Mikolov, Chen, Corrado, and
  Dean]{mikolov2013efficient}
Mikolov, T., Chen, K., Corrado, G., and Dean, J.
\newblock Efficient estimation of word representations in vector space.
\newblock \emph{arXiv preprint arXiv:1301.3781}, 2013.

\bibitem[Nam et~al.(2014)Nam, Kim, Menc{\'\i}a, Gurevych, and
  F{\"u}rnkranz]{nam2014large}
Nam, J., Kim, J., Menc{\'\i}a, E.~L., Gurevych, I., and F{\"u}rnkranz, J.
\newblock Large-scale multi-label text classification—revisiting neural
  networks.
\newblock In \emph{Joint european conference on machine learning and knowledge
  discovery in databases}, pp.\  437--452. Springer, 2014.

\bibitem[Nam et~al.(2017)Nam, Menc{\'\i}a, Kim, and
  F{\"u}rnkranz]{nam2017maximizing}
Nam, J., Menc{\'\i}a, E.~L., Kim, H.~J., and F{\"u}rnkranz, J.
\newblock Maximizing subset accuracy with recurrent neural networks in
  multi-label classification.
\newblock In \emph{Advances in neural information processing systems}, pp.\
  5413--5423, 2017.

\bibitem[Pennington et~al.(2014)Pennington, Socher, and
  Manning]{pennington2014glove}
Pennington, J., Socher, R., and Manning, C.~D.
\newblock Glove: Global vectors for word representation.
\newblock In \emph{Proceedings of the 2014 conference on empirical methods in
  natural language processing (EMNLP)}, pp.\  1532--1543, 2014.

\bibitem[Raykar et~al.(2010)Raykar, Yu, Zhao, Valadez, Florin, Bogoni, and
  Moy]{raykar2010learning}
Raykar, V.~C., Yu, S., Zhao, L.~H., Valadez, G.~H., Florin, C., Bogoni, L., and
  Moy, L.
\newblock Learning from crowds.
\newblock \emph{Journal of Machine Learning Research}, 11\penalty0 (4), 2010.

\bibitem[Read et~al.(2011)Read, Pfahringer, Holmes, and
  Frank]{read2011classifier}
Read, J., Pfahringer, B., Holmes, G., and Frank, E.
\newblock Classifier chains for multi-label classification.
\newblock \emph{Machine learning}, 85\penalty0 (3):\penalty0 333, 2011.

\bibitem[Russakovsky et~al.(2015)Russakovsky, Deng, Su, Krause, Satheesh, Ma,
  Huang, Karpathy, Khosla, Bernstein, et~al.]{russakovsky2015imagenet}
Russakovsky, O., Deng, J., Su, H., Krause, J., Satheesh, S., Ma, S., Huang, Z.,
  Karpathy, A., Khosla, A., Bernstein, M., et~al.
\newblock Imagenet large scale visual recognition challenge.
\newblock \emph{International journal of computer vision}, 115\penalty0
  (3):\penalty0 211--252, 2015.

\bibitem[Scarselli et~al.(2008)Scarselli, Gori, Tsoi, Hagenbuchner, and
  Monfardini]{scarselli2008graph}
Scarselli, F., Gori, M., Tsoi, A.~C., Hagenbuchner, M., and Monfardini, G.
\newblock The graph neural network model.
\newblock \emph{IEEE Transactions on Neural Networks}, 20\penalty0
  (1):\penalty0 61--80, 2008.

\bibitem[Snow et~al.(2008)Snow, O’connor, Jurafsky, and Ng]{snow2008cheap}
Snow, R., O’connor, B., Jurafsky, D., and Ng, A.~Y.
\newblock Cheap and fast--but is it good? evaluating non-expert annotations for
  natural language tasks.
\newblock In \emph{Proceedings of the 2008 conference on empirical methods in
  natural language processing}, pp.\  254--263, 2008.

\bibitem[Song et~al.(2020)Song, Kim, Park, and Lee]{song2020learning}
Song, H., Kim, M., Park, D., and Lee, J.-G.
\newblock Learning from noisy labels with deep neural networks: A survey, 2020.

\bibitem[Stock \& Cisse(2018)Stock and Cisse]{stock2018convnets}
Stock, P. and Cisse, M.
\newblock Convnets and imagenet beyond accuracy: Understanding mistakes and
  uncovering biases.
\newblock In \emph{Proceedings of the European Conference on Computer Vision
  (ECCV)}, pp.\  498--512, 2018.

\bibitem[Sullivan et~al.(2009)Sullivan, Wood, Iliff, Bonney, Fink, and
  Kelling]{sullivan2009ebird}
Sullivan, B.~L., Wood, C.~L., Iliff, M.~J., Bonney, R.~E., Fink, D., and
  Kelling, S.
\newblock ebird: A citizen-based bird observation network in the biological
  sciences.
\newblock \emph{Biological conservation}, 142\penalty0 (10):\penalty0
  2282--2292, 2009.

\bibitem[Sun et~al.(2019)Sun, Feng, Wang, Lang, and Jin]{sun2019partial}
Sun, L., Feng, S., Wang, T., Lang, C., and Jin, Y.
\newblock Partial multi-label learning by low-rank and sparse decomposition.
\newblock In \emph{Proceedings of the AAAI Conference on Artificial
  Intelligence}, volume~33, pp.\  5016--5023, 2019.

\bibitem[Tang et~al.(2018)Tang, Xue, Chen, and Gomes]{tang2018multi}
Tang, L., Xue, Y., Chen, D., and Gomes, C.~P.
\newblock Multi-entity dependence learning with rich context via conditional
  variational auto-encoder.
\newblock In \emph{AAAI}, 2018.

\bibitem[Vaswani et~al.(2017)Vaswani, Shazeer, Parmar, Uszkoreit, Jones, Gomez,
  Kaiser, and Polosukhin]{vaswani2017attention}
Vaswani, A., Shazeer, N., Parmar, N., Uszkoreit, J., Jones, L., Gomez, A.~N.,
  Kaiser, {\L}., and Polosukhin, I.
\newblock Attention is all you need.
\newblock In \emph{Advances in neural information processing systems}, pp.\
  5998--6008, 2017.

\bibitem[Veit et~al.(2017)Veit, Alldrin, Chechik, Krasin, Gupta, and
  Belongie]{veit2017learning}
Veit, A., Alldrin, N., Chechik, G., Krasin, I., Gupta, A., and Belongie, S.
\newblock Learning from noisy large-scale datasets with minimal supervision.
\newblock In \emph{Proceedings of the IEEE conference on computer vision and
  pattern recognition}, pp.\  839--847, 2017.

\bibitem[Wang et~al.(2018)Wang, Li, Wang, Zhang, Shen, Zhang, Henao, and
  Carin]{wang2018joint}
Wang, G., Li, C., Wang, W., Zhang, Y., Shen, D., Zhang, X., Henao, R., and
  Carin, L.
\newblock Joint embedding of words and labels for text classification.
\newblock In \emph{Proceedings of the 56th Annual Meeting of the Association
  for Computational Linguistics (Volume 1: Long Papers)}, pp.\  2321--2331,
  2018.

\bibitem[Wang et~al.(2016)Wang, Yang, Mao, Huang, Huang, and Xu]{wang2016cnn}
Wang, J., Yang, Y., Mao, J., Huang, Z., Huang, C., and Xu, W.
\newblock Cnn-rnn: A unified framework for multi-label image classification.
\newblock In \emph{Proceedings of the IEEE conference on computer vision and
  pattern recognition}, pp.\  2285--2294, 2016.

\bibitem[Wang et~al.()Wang, He, Li, Long, Zhou, Ma, and
  Wen]{DBLP:conf/aaai/WangHLLZMW20}
Wang, Y., He, D., Li, F., Long, X., Zhou, Z., Ma, J., and Wen, S.
\newblock Multi-label classification with label graph superimposing.
\newblock In \emph{The Thirty-Fourth {AAAI} Conference on Artificial
  Intelligence, New York, NY, USA, February 7-12, 2020}, pp.\  12265--12272.

\bibitem[Wang et~al.(2020)Wang, He, Li, Long, Zhou, Ma, and Wen]{wang2020multi}
Wang, Y., He, D., Li, F., Long, X., Zhou, Z., Ma, J., and Wen, S.
\newblock Multi-label classification with label graph superimposing.
\newblock In \emph{Proceedings of the AAAI Conference on Artificial
  Intelligence}, volume~34, pp.\  12265--12272, 2020.

\bibitem[Weston et~al.(2011)Weston, Bengio, and Usunier]{weston2011wsabie}
Weston, J., Bengio, S., and Usunier, N.
\newblock Wsabie: Scaling up to large vocabulary image annotation.
\newblock In \emph{Proceedings of the Twenty-Second International Joint
  Conference on Artificial Intelligence - Volume Volume Three}, IJCAI'11, pp.\
  2764–2770. AAAI Press, 2011.
\newblock ISBN 9781577355151.

\bibitem[Xie \& Huang(2018)Xie and Huang]{xie2018partial}
Xie, M.-K. and Huang, S.-J.
\newblock Partial multi-label learning.
\newblock In \emph{Thirty-Second AAAI Conference on Artificial Intelligence},
  2018.

\bibitem[Xie \& Huang(2020)Xie and Huang]{Xie_Huang_2020}
Xie, M.-K. and Huang, S.-J.
\newblock Partial multi-label learning with noisy label identification.
\newblock \emph{Proceedings of the AAAI Conference on Artificial Intelligence},
  34\penalty0 (04):\penalty0 6454--6461, Apr. 2020.
\newblock \doi{10.1609/aaai.v34i04.6117}.

\bibitem[Yao et~al.(2018)Yao, Wang, Tsang, Zhang, Sun, Zhang, and
  Zhang]{yao2018deep}
Yao, J., Wang, J., Tsang, I.~W., Zhang, Y., Sun, J., Zhang, C., and Zhang, R.
\newblock Deep learning from noisy image labels with quality embedding.
\newblock \emph{IEEE Transactions on Image Processing}, 28\penalty0
  (4):\penalty0 1909--1922, 2018.

\bibitem[You et~al.()You, Guo, Cui, Long, Bao, and
  Wen]{DBLP:conf/aaai/YouGCLBW20}
You, R., Guo, Z., Cui, L., Long, X., Bao, Y., and Wen, S.
\newblock Cross-modality attention with semantic graph embedding for
  multi-label classification.
\newblock In \emph{The Thirty-Fourth {AAAI} Conference on Artificial
  Intelligence, New York, NY, USA, February 7-12, 2020}, pp.\  12709--12716.

\bibitem[Yuen et~al.(2011)Yuen, King, and Leung]{yuen2011survey}
Yuen, M.-C., King, I., and Leung, K.-S.
\newblock A survey of crowdsourcing systems.
\newblock In \emph{2011 IEEE third international conference on privacy,
  security, risk and trust and 2011 IEEE third international conference on
  social computing}, pp.\  766--773. IEEE, 2011.

\bibitem[Zhang et~al.(2017{\natexlab{a}})Zhang, Bengio, Hardt, Recht, and
  Vinyals]{45820}
Zhang, C., Bengio, S., Hardt, M., Recht, B., and Vinyals, O.
\newblock Understanding deep learning requires rethinking generalization.
\newblock 2017{\natexlab{a}}.
\newblock URL \url{https://arxiv.org/abs/1611.03530}.

\bibitem[Zhang \& Fang(2020)Zhang and Fang]{zhang2020partial}
Zhang, M.-L. and Fang, J.-P.
\newblock Partial multi-label learning via credible label elicitation.
\newblock \emph{IEEE Transactions on Pattern Analysis and Machine
  Intelligence}, 2020.

\bibitem[Zhang \& Zhou(2007)Zhang and Zhou]{zhang2007ml}
Zhang, M.-L. and Zhou, Z.-H.
\newblock {ML-KNN}: A lazy learning approach to multi-label learning.
\newblock \emph{Pattern recognition}, 40\penalty0 (7):\penalty0 2038--2048,
  2007.

\bibitem[Zhang et~al.(2017{\natexlab{b}})Zhang, Yu, Kumar, and
  Chang]{zhang2017learning}
Zhang, X., Yu, F.~X., Kumar, S., and Chang, S.-F.
\newblock Learning spread-out local feature descriptors.
\newblock In \emph{Proceedings of the IEEE International Conference on Computer
  Vision}, pp.\  4595--4603, 2017{\natexlab{b}}.

\bibitem[Zhang \& Sabuncu(2018)Zhang and Sabuncu]{zhang2018generalized}
Zhang, Z. and Sabuncu, M.~R.
\newblock Generalized cross entropy loss for training deep neural networks with
  noisy labels.
\newblock In \emph{Proceedings of the 32nd International Conference on Neural
  Information Processing Systems}, pp.\  8792--8802, 2018.

\bibitem[Zhang et~al.(2020)Zhang, Zhang, Arik, Lee, and
  Pfister]{zhang2020distilling}
Zhang, Z., Zhang, H., Arik, S.~O., Lee, H., and Pfister, T.
\newblock Distilling effective supervision from severe label noise.
\newblock In \emph{Proceedings of the IEEE/CVF Conference on Computer Vision
  and Pattern Recognition}, pp.\  9294--9303, 2020.

\end{thebibliography}
